\documentclass[11pt,a4paper]{article}
\usepackage[hyperref]{naaclhlt2019}
\usepackage{times}
\usepackage{latexsym}

\usepackage{url}
\usepackage{graphicx}
\usepackage{paralist}
\usepackage{float}
\usepackage{amsfonts}
\usepackage{amsmath}
\usepackage{amsthm}
\usepackage{multirow}
\usepackage[para]{threeparttable}
\usepackage{xcolor}

\newcommand{\GRU}{\mathrm{GRU}}

\DeclareMathOperator{\maxpool}{maxpool}
\DeclareMathOperator{\meanpool}{meanpool}
\DeclareMathOperator{\softmax}{softmax}

\aclfinalcopy % Uncomment this line for the final submission
\def\aclpaperid{1093} %  Enter the acl Paper ID here

\title{PT-CoDE: Pre-trained Context-Dependent Encoder for \\Utterance-level Emotion Recognition}

\author{Wenxiang Jiao$^\dagger$, Michael R. Lyu$^\dagger$, {\rm and} Irwin King$^\dagger$ \\
  $^\dagger$~Department of Computer Science and Engineering, \\
  The Chinese University of Hong Kong, HKSAR, China \\
  {\tt \{wxjiao,lyu,king\}@cse.cuhk.edu.hk} \\}

\date{}

\begin{document}
\maketitle
\begin{abstract}
  Utterance-level emotion recognition (ULER) is a significant research topic for understanding human behaviors and developing empathetic chatting machines in the artificial intelligence area. Unlike traditional text classification problem, this task is supported by a limited number of datasets, among which most contain inadequate conversations or speeches. Such a data scarcity issue limits the possibility of training larger and more powerful models for this task. Witnessing the success of transfer learning in natural language process (NLP), we propose to pre-train a context-dependent encoder (CoDE) for ULER by learning from unlabeled conversation data. Essentially, CoDE is a hierarchical architecture that contains an utterance encoder and a conversation encoder, making it different from those works that aim to pre-train a universal sentence encoder. Also, we propose a new pre-training task named ``conversation completion" (CoCo), which attempts to select the correct answer from candidate answers to fill a masked utterance in a question conversation. The CoCo task is carried out on pure movie subtitles so that our CoDE can be pre-trained in an unsupervised fashion. Finally, the pre-trained CoDE (PT-CoDE) is fine-tuned for ULER and boosts the model performance significantly on five datasets.
\end{abstract}

\section{Introduction}
\label{sec:introduction}

Sentiment analysis is a prevalent topic in natural language processing (NLP) area. It has been widely applied in scenarios that involve the voice of customers, such as reviews and survey responses in marketing, and social media in customer service. We consider one of the tasks in this research direction, known as utterance-level emotion recognition (ULER)~\cite{DBLP:conf/acl/PoriaCHMZM17}, which needs to identify the emotion state of each utterance in a conversation or speech. This task is significant for understanding human behaviors in conversations or speeches and developing empathetic chatting machines.

There are two characteristics of ULER that distinguish it from traditional sentiment analysis. On one hand, the instances are utterances from a conversation or speech, which indicates an inherent hierarchical structure of data, i.e., words-to-utterance and utterances-to-conversation/speech. The sequential relationship between utterances suggests the need of extracting contextual information for each utterance. As a result, to tackle this task, current approaches~\cite{DBLP:conf/acl/PoriaCHMZM17,DBLP:conf/naacl/HazarikaPZCMZ18} adopt context-dependent models with a recurrent neural network (RNN) to read the utterance embeddings and then produce the contextual representations for prediction. 
On the other hand, in ULER, each instance should be tagged with a specific emotion label, rather than coarse polarities such as positive and negative. This requires annotators to recognize either obvious or subtle difference between emotions, making it infeasible to label data manually in large scale. In fact, there are a very limited number of datasets for ULER, among which most contain few conversations or speeches. So far, the largest dataset for ULER is MOSEI~\cite{DBLP:conf/acl/MorencyCPLZ18}, with {23,543} utterances in {3,228} speeches. Other datasets have even less conversations (IEMOCAP~\cite{DBLP:journals/lre/BussoBLKMKCLN08}: 151, Friends~\cite{DBLP:conf/acl-socialnlp/HsuK18}: {1,000},
EmotionPush~\cite{DBLP:conf/acl-socialnlp/HsuK18}: {1,000}, and EmoryNLP~\cite{DBLP:conf/aaai/ZahiriC18}: {897}), which are inadequate for sufficient training of the context-dependent models.

Based on these two characteristics, we propose PT-CoDE, i.e., a Pre-trained Context-Dependent Encoder, for the ULER task. Our contributions are summarized as follows: (1) We propose to pre-train a Context-Dependent Encoder (CoDE) to boost its performance in ULER. Essentially, our CoDE consists of two parts: an utterance encoder to extract the context-independent embeddings for utterances, and a conversation encoder to capture the utterance-level context. Such a hierarchical architecture makes our work different from those that aim to pre-train a universal sentence encoder~\cite{DBLP:conf/acl/RuderH18,radford2018improving,DBLP:conf/naacl/DevlinCLT19}.
(2) To enable the pre-training of CoDE in an unsupervised fashion, we propose a new task called ``conversation completion" (CoCo), which attempts to select the correct answer from candidate answers to fill a masked utterance in a question conversation. The candidate answers are composed by the original utterance before masked and some noise utterances randomly sampled elsewhere. We carry out the CoCo task on pure movie subtitles, which are usually conversations between characters. The task itself is also worth further investigation in NLP, since in real life it helps people to enhance reading and understanding skills. 
(3) We transfer and fine-tune our PT-CoDE to complete the ULER task and achieve significant improvement of model performance on five datasets.

% Related work
\section{Related Work}
\label{sec:relatedwork}

With the thrive of deep learning techniques, many prominent approaches for text classification has been explored, including recursive autoencoders (RAEs)~\cite{DBLP:conf/emnlp/SocherPHNM11}, convolutional neural networks (CNNs)~\cite{DBLP:conf/emnlp/Kim14}, recurrent neural networks (RNNs)~\cite{DBLP:conf/acl/Abdul-MageedU17}, and dynamic memory networks (DMNs)~\cite{DBLP:conf/icml/KumarIOIBGZPS16}. These approaches focus on extracting high-quality representations for independent text instances without sentence-level context. Thus they cannot sufficiently exploit the contextual information in ULER. To take advantage of context, \cite{DBLP:conf/acl/PoriaCHMZM17} propose a bidirectional contextual long short-term memory (LSTM) network, termed as bcLSTM, to capture the sequential relationship of utterances. \cite{DBLP:conf/emnlp/HazarikaPMCZ18} and \cite{DBLP:conf/naacl/HazarikaPZCMZ18} propose conversational memory networks (CMNs) for dyadic conversation scenarios and achieve better performance than bcLSTM. \cite{DBLP:conf/acl-socialnlp/LuoYC18} adopt a self-attention mechanism to capture the context, even though the model does not work very well and is biased to majority classes. These context-dependent models are well motivated, but the problem is that there is no enough labeled data to sufficiently train these models. 

Pre-training on unlabeled data has been an active area of research for decades. \cite{DBLP:conf/nips/MikolovSCCD13} and \cite{DBLP:conf/emnlp/PenningtonSM14} lead the heat on learning dense word embeddings over raw text for downstream tasks. \cite{DBLP:conf/conll/MelamudGD16} propose to learn word embeddings in the context with the use of LSTM, which is able to eliminate word-sense ambiguity. More recently, ELMo~\cite{DBLP:conf/naacl/PetersNIGCLZ18} extracts context-sensitive features through a language model and integrates the features into task-specific architectures, achieving state-of-the-art results on several major NLP tasks. Unlike these feature-based approaches, another trend is to pre-train some architecture through a language model objective, and then fine-tune the architecture for supervised downstream tasks~\cite{DBLP:conf/acl/RuderH18,radford2018improving,DBLP:conf/naacl/DevlinCLT19}. With trainable parameters, this kind of approaches are more flexible, attaining better performance than their feature-based counterparts. 

However, the idea of pre-training a CoDE using unlabeled conversation data for ULER has never been explored. On one hand, existing work on ULER mentioned above focus on modeling the speakers, context, and emotion evolution. No prior work has tried to solve the issue of data scarcity. On the other hand, existing work on transfer learning focuses on pre-training universal sentence encoders, e.g., ELMo, GPT, and BERT. But our PT-CoDE, beyond sentence level, is dedicated for sentence sequences from conversations or speeches.
As a result, the pre-training task needs to be customized, for which we propose the CoCo task. Partially inspired by Word2vec~\cite{DBLP:conf/nips/MikolovSCCD13} and response selection task~\cite{DBLP:conf/acl/TongZJM17}, our CoCo task differs in that it should model the order of context meanwhile both historical and future context are provided. In contrast, Word2vec neglects the order of context words, and response selection task usually provides only historical context. 

% PCUE: Pre-trained Contextual Utterance Encoder
\section{PT-CoDE}
\label{sec:PT-CoDE}

In this section, we introduce the pre-training details of CoDE. We will cover
the architecture of CoDE, the CoCo task, the dataset creation, and the pre-training procedures.

% CoDE
\subsection{CoDE: Context-Dependent Encoder}
\label{ssec:CoDE}

A classical context-dependent encoder includes three main components, namely, a word embedding layer, an utterance encoder, and a conversation encoder. It simulates the hierarchical structure of a conversation that contains words-to-utterance and utterances-to-conversation relationships.
\\[1ex]
\noindent\textbf{Word Embedding Layer.}
For an utterance with a sequence of words denoted by indices in the vocabulary $W=\{w_1, w_2, \cdots, w_T\}$, the word embedding layer will output a vector for each word $\mathbf{X} = \{ \mathbf{x}_1, \mathbf{x}_2, \cdots, \mathbf{x}_T \}$, where $\mathbf{x}_i \in \mathbb{R}^{d_w}$, and $T$ is the length of the sequence.
Here, we utilize the 300-dimensional pre-trained GloVe word vectors\footnote{https://nlp.stanford.edu/projects/glove/}~\cite{DBLP:conf/emnlp/PenningtonSM14} trained over 840B Common Crawl to initialize the word embedding layer. Those words that cannot be found in the GloVe vocabulary are initialized by randomly generated vectors.
\\[1ex]
\noindent\textbf{Utterance Encoder.}
We choose a bidirectional gated recurrent unit~\cite{DBLP:conf/emnlp/ChoMGBBSB14}, termed as BiGRU, to model the word sequence of an utterance. The BiGRU takes the word embeddings $\mathbf{X} \in \mathbb{R}^{T \times d_w}$ of the utterance as input and produces the contextual word embeddings, i.e., the hidden states $\overrightarrow{\mathbf{h}}_t, \overleftarrow{\mathbf{h}}_t \in \mathbb{R}^{d_u}$. We apply \textit{max-pooling} and \textit{mean-pooling} on the hidden states and sum up the pooling results, followed by a fully-connected (FC) layer, to obtain the embedding of an utterance $\mathbf{u}_l$. The steps above are formulated as:
\begin{align}
\overrightarrow{\mathbf{h}}_t &= \overrightarrow{\GRU}(\mathbf{x}_t, \overrightarrow{\mathbf{h}}_{t-1}), t \in [1, T], \\
\overleftarrow{\mathbf{h}}_t &= \overleftarrow{\GRU}(\mathbf{x}_t, \overleftarrow{\mathbf{h}}_{t+1}), t \in [1, T], \\
\mathbf{h}_l &= \maxpool(\{[\overrightarrow{\mathbf{h}}_t; \overleftarrow{\mathbf{h}}_t]\}_{t=1}^T) \\
&+ \meanpool(\{[\overrightarrow{\mathbf{h}}_t; \overleftarrow{\mathbf{h}}_t]\}_{t=1}^T), \\
\mathbf{u}_l &= \tanh(\mathbf{W}_u \cdot \mathbf{h}_l + \mathbf{b}_u), l \in [1, L],
\end{align}
where $\overrightarrow{\GRU}$ and $\overleftarrow{\GRU}$ represent the forward and backward GRUs, respectively, $\mathbf{W}_u \in \mathbb{R}^{d_u \times 2d_u}, \mathbf{b}_u \in \mathbb{R}^{d_u}$ are trainable weights, and $L$ is the length of a conversation.
\\[1ex]
\noindent\textbf{Conversation Encoder.}
Like the dependency between a word and its context, an utterance could express different meanings given different context. To capture the relationship between the utterances, we adopt another BiGRU to model the utterance sequence of a conversation:
\begin{align}
\overrightarrow{\mathbf{H}}_l &= \overrightarrow{\GRU}(\mathbf{u}_l, \overrightarrow{\mathbf{H}}_{l-1}), l \in [1, L], \\
\overleftarrow{\mathbf{H}}_l &= \overleftarrow{\GRU}(\mathbf{u}_l, \overleftarrow{\mathbf{H}}_{l+1}), l \in [1, L].
\end{align}
where $\overrightarrow{\mathbf{H}}_l, \overleftarrow{\mathbf{H}}_l \in \mathbb{R}^{d_c}$ are the hidden states.

%fig:CoCo
\begin{figure*}[t!]
\parbox{0.36\textwidth}{
    \centering
    \includegraphics[width=0.9\columnwidth]{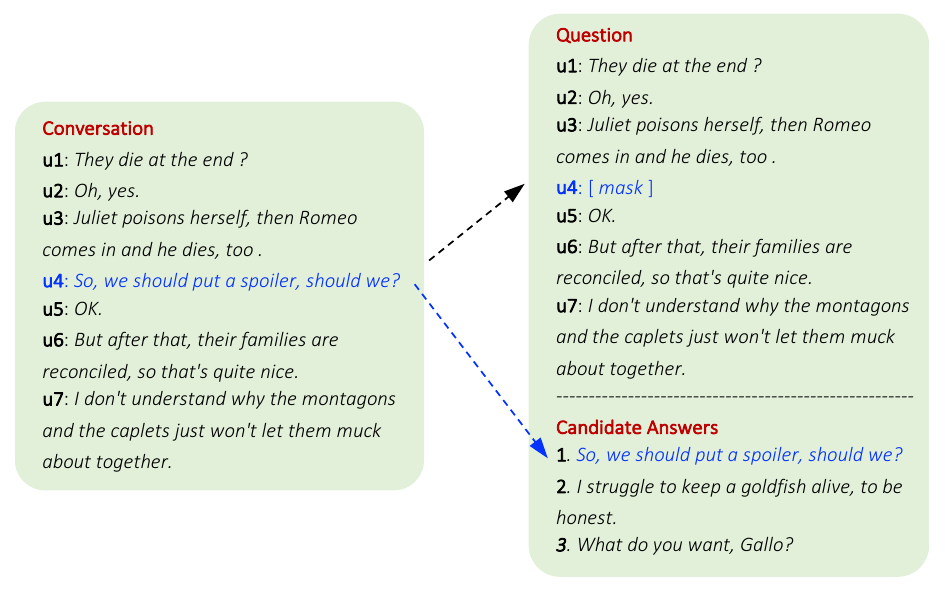}
    \caption{A data example in the CoCo task.} 
    \label{fig:CoCo}
}
\hfill
\parbox{0.6\textwidth}{
    \centering
    \includegraphics[width=1.1\columnwidth]{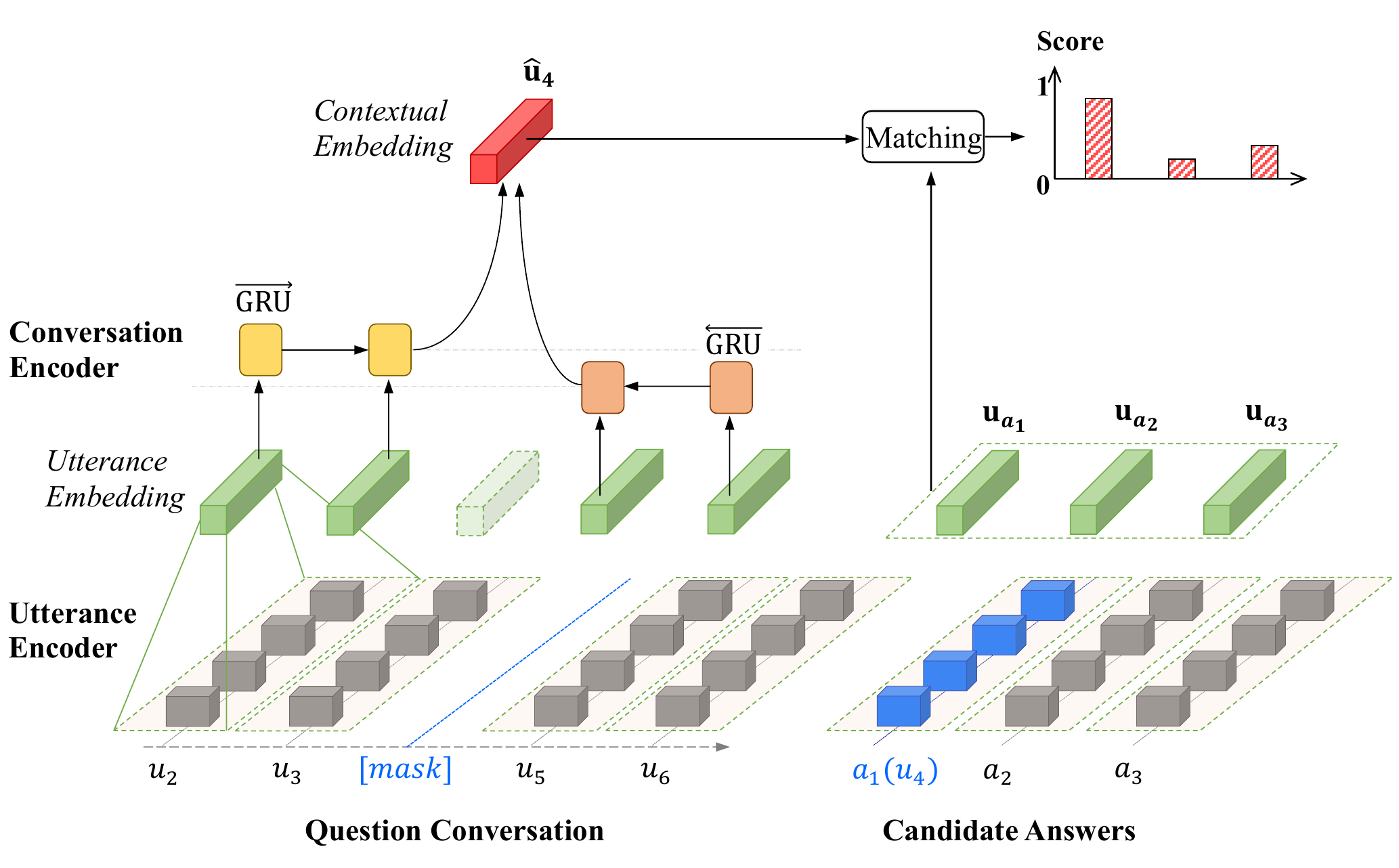}
    \caption{The architecture of CoDE with the pre-training objective.}
\label{fig:CoDE}
}
\end{figure*}

% Pre-training Task
\subsection{Pre-training Task}
\label{ssec:pretrain_task}

Language modeling  (LM)~\cite{DBLP:journals/corr/JozefowiczVSSW16,DBLP:conf/iclr/MelisDB18} is the most prevalent way to pre-train universal sentence encoders, which make us wonder whether our CoDE can be pre-trained in a similar way. We know that the goal of LM is to predict the current word given the historical words. By putting an utterance in the position of a word, we can define a conversation model (CM).
Suppose we have a conversation with $L$ utterances, i.e., $\mathcal{U}=\{ u_1, u_2, \cdots, u_L \}$, a CM computes the probability of this conversation by modeling the probability of utterance $l$ given the history $(u_1, u_2, \cdots, u_{l-1})$:
\begin{align}
p(u_1, u_2, \cdots, u_L) = \prod_{l=1}^{l=L}p(u_l|u_1, u_2, \cdots, u_{l-1}).
\end{align}

Similar to LM, we can compute a context-independent utterance embedding $\mathbf{u}_l$ and then pass it to forward RNNs, such as LSTM and GRU. For each position $l$, the output of RNN is used to predict the next utterance $u_{l+1}$ with a softmax layer.
The idea of utilizing conversation models for pre-training, just as LM, is conceptually simple. However, the size of the sentence classes is much larger than the vocabulary in LM, making the softmax function intractable. Meanwhile, the number of sentences in each class could be very small, which is inadequate for model training.
Thus, we propose the CoCo task as an alternative.
\\[1ex]
\noindent\textbf{CoCo: Conversation Completion.}
To enable the pre-training of CoDE on pure conversations, we exploit the self-supervision information in conversations, i.e., the ordering information of utterances. We obtain the inspiration from Word2vec~\cite{DBLP:conf/nips/MikolovSCCD13} and response selection task~\cite{DBLP:conf/acl/TongZJM17} to predict a target utterance replying on the other utterances around it, i.e., the context utterances. 

Formally, our CoCo task is defined as:
For a given conversation, $\mathcal{U}=\{ u_1, u_2, \cdots, u_L \}$, we mask a target utterance $u_l$ as $\mathcal{U}\backslash{u_l}=\{ \cdots, u_{l-1}, [mask] , u_{l+1}, \cdots \}$ to create a question. Subsequently, we put the target utterance together with $N-1$ noise utterances to form a set of $N$ candidate answers. For example, in Figure~\ref{fig:CoCo}, the utterance \texttt{u4} is masked out from the original conversation and \texttt{two} noise utterances are sampled elsewhere together with \texttt{u4} to form the candidate answers. The goal of our CoCo task is to select the correct answer, i.e., $u_l$, from the candidate answers to fill the mask, conditioned on the context utterances.
\\[1ex]
\textbf{Pre-training Objective.}
To adapt CoDE for the proposed CoCo task, we construct a contextual embedding for each masked utterance by combining its context from the history $\overrightarrow{\mathbf{H}}_{l-1}$ and the future $\overleftarrow{\mathbf{H}}_{l+1}$ (see Figure~\ref{fig:CoDE}):
\begin{align}
\mathbf{\hat{u}}_l = \tanh(\mathbf{W}_c \cdot [\overrightarrow{\mathbf{H}}_{l-1}; \overleftarrow{\mathbf{H}}_{l+1}] + \mathbf{b}_c) ,
\end{align}
where  $\mathbf{W}_c \in \mathbb{R}^{d_u \times 2d_c}, \mathbf{b}_c \in \mathbb{R}^{d_u}$ are trainable weights.

The contextual embedding $\mathbf{\hat{u}}_l$ is matched to the candidate answers to find the most suitable one to fill the mask. To compute the matching score, we adopt dot-product with a sigmoid function as:   
\begin{align}
s(\mathbf{\hat{u}}_l, \mathbf{u}_{a_n}) = \sigma(\mathbf{\hat{u}}_l ^\top \mathbf{u}_{a_n}), n \in [1, N],
\end{align}
where $\sigma(x)=\frac{1}{(1+\exp(-x))} \in (0, 1)$ is the sigmoid function, and $\mathbf{u}_{a_n}$ is the embedding of the $n$th candidate answer.

The higher the matching score is, the better the candidate answer fills the mask. Therefore, we need to maximize the score of the target utterance and minimize the score of the noise utterances. The loss function becomes:
\begin{align}
\mathcal{F} = -\sum_{l}\left[ \log\sigma(\mathbf{\hat{u}}_l ^\top \mathbf{u}_{a_1}) + \sum_{n=2}^{N}\log\sigma(-\mathbf{\hat{u}}_l ^\top \mathbf{u}_{a_n}) \right],
\end{align}
where $a_1$ corresponds to the target utterance, and the summation goes over each utterance of all the conversations.

\iffalse
%fig:CoDE
\begin{figure}[t!]
\centering
\includegraphics[width=1\columnwidth]{CoDE.pdf}
\caption{The architecture of CoDE with the pre-training objective.}
\label{fig:CoDE}
\end{figure}
\fi

% CoCo Dataset
\subsection{Dataset Creation}
\label{ssec:CoCoData}

To rely on the CoCo task for pre-training, we need a large amount of conversation data. OpenSubtitle\footnote{http://opus.nlpl.eu/OpenSubtitles-v2018.php}~\cite{DBLP:conf/lrec/LisonT16} is a very large database of movie and TV show subtitles, which are usually conversations between characters. We retrieve the English subtitles of movies and TV shows throughout the year of 2016, including {25,466} \texttt{.html} files, to create our dataset for pre-training.
\\[1ex]
\textbf{Preprocessing.}
We extract the text subtitles from all the \texttt{.html} files and preprocess them as below:

(1) For each episode, we remove the first and the last \textit{ten} utterances in case they are instructions but conversations, especially in TV shows.

(2) We split the conversations in each episode randomly into shorter ones with \textit{five} to \textit{one hundred} utterances, following a uniform distribution.

(3) A short conversation is removed if over half of its utterances contain less than \textit{eight} words each. This is done to force the conversation to capture more information.

(4) All the short conversations are randomly split into a training set, a validation set, and a testing set, following the ratio of 90:5:5.

The statistics of resulted sets are presented in Table~\ref{table:StatCCD}, where {\bf \#Conversation} denotes the number of conversations in a set, {\bf AvgU} is the average number of utterances in a conversation, and {\bf AvgW} is the average number of tokens in an utterance. Totally, there are over 2 million of utterances in over 60k conversations, which is at least 100 times more than those datasets for ULER (see Table~\ref{table:StatULER}).
\\[1ex]
\textbf{Noise Utterances.}
We randomly sample \textit{ten} noise utterances for each utterance in the training set, validation set, and testing set. In each set, a conversation shares the \textit{ten} noise utterances sampled from elsewhere within the set. During training, we can either use the pre-selected noise utterances or sample an arbitrary number of noise utterances dynamically. We use the validation set to choose model parameters, and test the model performance on the testing set.

% Pre-training & Evaluation
\subsection{Pre-training \& Evaluation}
\label{ssec:Pretraining}

In this section, we introduce the pre-training procedure and evaluation of CoDE.
\\[1ex]
\textbf{Evaluation Metric.}
In our CoCo task, we adopt the evaluation metric $\mathbf{R}_{N'}@k = \frac{\sum_{i=1}^{k}y_i}{\sum_{i=1}^{N'}y_i}$~\cite{DBLP:conf/acl/WuLCZDYZL18}, which is the recall of the true positives among $k$ best-matched answers from $N'$ available candidates for the given contextual embedding $\mathbf{\hat{u}}_k$. The variate $y_i$ represents the binary label for each candidate, i.e., $1$ for the target one and $0$ for the noise ones. Here, we report $\mathbf{R}_{5}@1$, $\mathbf{R}_{5}@2$, $\mathbf{R}_{11}@1$, and $\mathbf{R}_{11}@2$.
\\[1ex]
\textbf{Pre-training Procedures.}
We train our CoDE on the created dataset in three scales:
\begin{itemize}
\item[-] PT-CoDE$_{small}$: $d_u = 150$, $d_c = 150$;
\item[-] PT-CoDE$_{mid}$: $d_u = 300$, $d_c = 300$;
\item[-] PT-CoDE$_{large}$: $d_u = 450$, $d_c = 450$.
\end{itemize}

We choose Adam~\cite{DBLP:journals/corr/KingmaB14} as the optimizer with an initial learning rate of $2\times10^{-4}$, which is decayed with a rate of 0.75 once the validation recall $\mathbf{R}_{11}@1$ stops increasing. We use a dropout rate of 0.5 for the utterance encoder and the conversation encoder, respectively. Gradient clipping with a norm of 5 is also applied to avoid gradient explosion.
Each conversation in the training set is regarded as a batch, where each utterance plays the role of target utterance by turns. We randomly sample 10 noise utterances for each conversation during training and validate the model every epoch. The CoDE is pre-trained for at most 20 epochs, and early stopping with a patience of 3 is adopted to choose the optimal parameters. Note that, we fix the word embedding layer during pre-training to focus on the utterance encoder and the conversation encoder.
\\[1ex]
\textbf{Testing Results.}
The results on the testing set are shown in Table~\ref{table:TestPT-CoDE}. 
For the small model PT-CoDE$_{small}$, it is able to select the correct answer for 70.8\% instances with 5 candidate answers and 56.2\% with 11 candidates. The accuracy is considerably higher than random guesses, i.e., 1/5 and 1/11, respectively. By increasing the model scale to PT-CoDE$_{mid}$ and PT-CoDE$_{large}$, we further improve the recalls by several points successively. These results demonstrate that CoDE is indeed able to capture the structure of conversations and perform well in the proposed conversation completion task.

% Statistics of the conversation completion dataset
\begin{table}[t!]
\small
\centering
\begin{threeparttable}
%\resizebox{\columnwidth}{!}{
\begin{tabular}{|l|ccc|}
\hline
{\bf Set} & {\bf \#Conversation} & {\bf AvgU} & { \bf AvgW} \\
\hline
Train & {58,360} & 41.3 & 10.1  \\
Val & {3,186} & 41.0 & 10.1 \\
Test & {3,297} & 40.8 & 10.1 \\
\hline
\end{tabular}
%}
\end{threeparttable}
\caption{Statistics of the created dataset for the CoCo task.}
\label{table:StatCCD}
\end{table}

% Results of PT-CoDE
\begin{table}[t!]
\small
\centering
\begin{threeparttable}
%\resizebox{0.98\columnwidth}{!}{
\begin{tabular}{|l|cccc|}
\hline
\textbf{Model} & $\mathbf{R}_{5}@1$ & $\mathbf{R}_{5}@2$ & $\mathbf{R}_{11}@1$ & $\mathbf{R}_{11}@2$ \\
\hline
PT-CoDE$_{small}$ & 70.8 & 88.0 & 56.2 & 72.7 \\
PT-CoDE$_{mid}$ & 73.8 & 89.7 & 60.4 & 76.4 \\
PT-CoDE$_{large}$ & 77.2 & 91.3 & 64.2 & 79.1 \\
\hline
\end{tabular}
%}
\end{threeparttable}
\caption{Testing results of CoDE on the CoCo task.}
\label{table:TestPT-CoDE}
\end{table}

% PT-CoDE for Utterance-level Emotion Recognition
\section{Fine-tuning PT-CoDE for ULER}
\label{sec:ULER}

In this section, we present how to fine-tune PT-CoDE for the ULER task. We first introduce the architecture for ULER transferred from PT-CoDE, then the datasets, the evaluation metrics, and the compared methods. Finally, we report our experimental results along with extensive analysis.

\subsection{ULER Architecture}

To transfer PT-CoDE to the ULER task, we only need to add a fully-connected (FC) layer followed by a softmax function to form the new architecture. Here, we also concatenate the context-independent utterance embeddings to the contextual ones before fed to the FC. The formulations are as below:
\begin{align}
\mathbf{H}_l &= \tanh(\mathbf{W}_c' \cdot [\overrightarrow{\mathbf{H}}_l; \overleftarrow{\mathbf{H}}_l;\mathbf{u}_l] + \mathbf{b}_c') , \\
\mathbf{o}_l &= \softmax(\mathbf{W}_f \cdot \mathbf{H}_l + \mathbf{b}_f ),
\end{align}
where $\mathbf{W}_c' \in \mathbb{R}^{d_f \times (2d_c+d_u)}$, $\mathbf{b}_c' \in \mathbb{R}^{d_f}$, $\mathbf{W}_f \in \mathbb{R}^{|\mathcal{C}| \times d_f}$, $\mathbf{b}_f \in \mathbb{R}^{|\mathcal{C}|}$ are trainable parameters. $d_f$ is the dimension of FC, and $|\mathcal{C}|$ is the number of emotion classes.

We adopt a weighted categorical cross-entropy loss function to optimize the model parameters:
\begin{align}
\mathcal{L} = -\frac{1}{\sum_{i=1}^{N}L_i}\sum_{i=1}^{N} \sum_{j=1}^{L_i} \omega(c_j) \sum_{c=1}^{|\mathcal{C}|} \mathbf{o}_j^c\log_2(\mathbf{\hat{o}}_j^c),
\end{align}
where $\mathbf{o}_j$ is the one-hot vector of the true label, and $\mathbf{o}_j^c$ and $\mathbf{\hat{o}}_j^c$ are the elements of $\mathbf{o}_j$ and $\mathbf{\hat{o}}_j$ for class $c$. The weight $\omega(c)$ is inversely proportional to the percentage of class $c$ in the training set with a power rate of 0.5.

% Baselines
\subsection{Baseline Methods}
\label{ssec:baseline}

We mainly compare our PT-CoDE$_{mid}$ with bcLSTM~\cite{DBLP:conf/acl/PoriaCHMZM17}, CMN~\cite{DBLP:conf/naacl/HazarikaPZCMZ18}, 
SA-BiLSTM~\cite{DBLP:conf/acl-socialnlp/LuoYC18}, CNN-DCNN~\cite{DBLP:conf/acl-socialnlp/Khosla18},
SCNN~\cite{DBLP:conf/aaai/ZahiriC18}, and the following ones implemented by ourselves:

$\bullet$~bcLSTM$_{\ddagger}$~\cite{DBLP:conf/acl/PoriaCHMZM17}: A bcLSTM with a 1-D CNN to extract the utterance embeddings, and a bidirectional LSTM to model the relationship of utterances. The 1-D CNN has filter sizes of 2, 3, and 4 with 100 feature maps each, and the size of hidden state of the LSTM is 300.

$\bullet$~bcGRU: A variant of bcLSTM$_{\ddagger}$ with a BiGRU to capture the utterance-level context.

$\bullet$~CoDE$_{mid}$: The middle scale CoDE proposed in this paper without pre-training.

% Data Statistics
\begin{table}[t!]
\small
\centering
\begin{threeparttable}
\resizebox{0.98\columnwidth}{!}{
\begin{tabular}{|l|ccc|ccc|}
\hline
\multirow{2}{*}{\bf Model}
& \multicolumn{3}{c|}{\bf \#Conversation}
& \multicolumn{3}{c|}{\bf \#Utterance} \\
\cline{2-7}
& \bf Train & \bf Val & \bf Test & \bf Train & \bf Val & \bf Test  \\
\hline
IEMOCAP & 96 & 24 & 31 & {3,569} & 721 & {1,208} \\
Friends & 720 & 80 & 200 & {10,561} & {1,178} & {2,764} \\
EmotionPush & 720 & 80 & 200 & {10,733} & {1,202} & {2,807} \\
EmoryNLP & 713 & 99 & 85 & {9,934} & {1,344} & {1,328} \\
MOSEI$^*$ & {2,250} & 300 & 676 & {16,331} & {1,871} & {4,662} \\
\hline
\end{tabular}
}
\end{threeparttable}
\caption{Statistics of the datasets for ULER.}
\label{table:StatULER}
\end{table}

\subsection{ULER Datasets}
\label{ssec:ULERDataset}

Our PT-CoDE$_{mid}$ and the implemented baselines are tested on five ULER datasets, namely, 
IEMOCAP\footnote{https://sail.usc.edu/iemocap/}~\cite{DBLP:journals/lre/BussoBLKMKCLN08}, Friends\footnote{http://doraemon.iis.sinica.edu.tw/emotionlines}~\cite{DBLP:conf/lrec/HsuCKHK18}, 
EmotionPush\footnote{http://doraemon.iis.sinica.edu.tw/emotionlines}~\cite{DBLP:conf/lrec/HsuCKHK18}, 
EmoryNLP\footnote{https://github.com/emorynlp/emotion-detection/}~\cite{DBLP:conf/aaai/ZahiriC18}, 
and MOSEI\footnote{http://immortal.multicomp.cs.cmu.edu/raw\_datasets/}~\cite{DBLP:conf/acl/MorencyCPLZ18}. 
For MOSEI, we utilize the raw transcripts, where over 14k utterances are not annotated, and others are labeled with one or more emotion labels.
For the unlabeled utterances, we just remove them from the dataset. For the utterance with more than one emotion labels, we determine its primary emotion by the majority vote or the highest emotion intensity sum if there are more than one majority votes. For the utterances that obtain zero vote for all emotion classes, we annotate them as \textit{other}. The resulted dataset is different from the official one, so we name it MOSEI$^*$ here.

For the first three datasets, we follow previous work~\cite{DBLP:conf/acl/PoriaCHMZM17,DBLP:conf/lrec/HsuCKHK18} to consider only four emotion classes, i.e., \textit{anger}, \textit{joy}, \textit{sadness}, and \textit{neutral}. We consider all the emotion classes for EmoryNLP as in \cite{DBLP:conf/aaai/ZahiriC18} and six emotion classes (without \textit{neutral}) for MOSEI$^*$.
All the datasets contain the training set, validation set, and testing set, except for IEMOCAP. So, we follow \cite{DBLP:conf/acl/PoriaCHMZM17} to use the first four sessions of transcripts as the training set, and the last one as the testing set. The validation set is extracted from the randomly-shuffled training set with the ratio of 80:20. We present the statistic details of datasets in Table~\ref{table:StatULER}.

%%%%%% EmoryNLP and MOSEI
\begin{table*}[t!]
\small
\centering
%\resizebox{0.92\textwidth}{!}{
\begin{threeparttable}
\begin{tabular}{|l|ccc|ccc|ccc|}
\hline
\multirow{2}{*}{\bf Model}
& \multicolumn{3}{c|}{\bf IEMOCAP}
& \multicolumn{3}{c|}{\bf EmoryNLP}
& \multicolumn{3}{c|}{\bf MOSEI$^*$} \\
\cline{2-10}
& {\bf F1} & {\bf WA} & {\bf UA} & {\bf F1} & {\bf WA} & {\bf UA} & {\bf F1} & {\bf WA} & {\bf UA} \\
\hline
bcLSTM-uni/tri\tnote{1} &-/- & 73.6/76.1 & 74.6/76.3 &-&-&- &-&-&-\\
CMN-uni/tri\tnote{2} &-/- & 74.1/77.6 & -/79.1 &-&-&- &-&-&- \\
SCNN\tnote{3} &-&-&-& 26.9 & \bf 37.9 & - & - & - & - \\
\hline
bcLSTM$_\ddagger$ & 76.6$_{\pm 1.7}$ & 77.1$_{\pm 1.6}$ & 78.0$_{\pm 1.4}$ & 25.5$_{\pm 0.8}$ & 33.5$_{\pm 1.8}$ & 27.6$_{\pm 0.8}$ & 29.1$_{\pm 0.5}$ & 56.3$_{\pm 1.7}$ & 29.8$_{\pm 0.8}$ \\
bcGRU & 77.6$_{\pm 2.1}$ & 78.2$_{\pm 2.2}$ & 78.7$_{\pm 1.1}$ & 26.1$_{\pm 0.7}$ & 33.1$_{\pm 1.6}$ & 27.4$_{\pm 0.4}$ & 28.7$_{\pm 0.8}$ & 56.4$_{\pm 1.8}$ & 29.8$_{\pm 1.1}$ \\
CoDE$_{mid}$ & 78.6$_{\pm 1.3}$ & 79.6$_{\pm 1.7}$ & 78.9$_{\pm 1.3}$ & 26.7$_{\pm 0.3}$ & 34.7$_{\pm 1.4}$ & 28.8$_{\pm 0.4}$ & 29.7$_{\pm 0.7}$ & 56.6$_{\pm 1.2}$ &  30.6$_{\pm 0.7}$ \\
\hline
PT-CoDE$_{mid}$ & \bf 81.5$_{\pm 0.4}$ & \bf 82.9$_{\pm 0.3}$ & \bf 81.7$_{\pm 0.4}$ & \bf 29.1$_{\pm 0.6}$ & 36.1$_{\pm 1.3}$ & \bf 30.3$_{\pm 0.9}$ & \bf 31.7$_{\pm 0.2}$ & \bf 57.1$_{\pm 2.0}$ & \bf 33.0$_{\pm 1.1}$ \\
\hline
\end{tabular}
\begin{tablenotes}\footnotesize
\item[1]by~\cite{DBLP:conf/acl/PoriaCHMZM17};
\item[2]by~\cite{DBLP:conf/naacl/HazarikaPZCMZ18};
\item[3]by~\cite{DBLP:conf/aaai/ZahiriC18}.
\end{tablenotes}
\end{threeparttable}
%}
\caption{Testing results on IEMOCAP, EmoryNLP, and MOSEI$^*$.}
\label{table:res_IEMOCAP_EmoryNLP_MOSEI}
\end{table*}

%%%%% Friends and EmotionPush
\begin{table*}[t!]
\parbox{0.62\textwidth}{
\centering
\begin{threeparttable}
\resizebox{0.6\textwidth}{!}{
\begin{tabular}{|l|ccc|ccc|}
\hline
\multirow{2}{*}{\bf Model}
& \multicolumn{3}{c|}{\bf Friends}
& \multicolumn{3}{c|}{\bf EmotionPush} \\
\cline{2-7}
& {\bf F1} & {\bf WA} & {\bf UA} & {\bf F1} & {\bf WA} & {\bf UA} \\
\hline
SA-BiLSTM\tnote{1} &- & 79.8 & 59.6 &- & \bf 87.7 & 55.0 \\
CNN-DCNN\tnote{2} &- & 67.0 & 62.5 &- & 75.7 & \bf 62.5 \\
\hline
bcLSTM$_\ddagger$ & 63.1$_{\pm 0.6}$ & 79.9$_{\pm 0.5}$ & 63.3$_{\pm 1.5}$ & 60.3$_{\pm 0.7}$ & 84.8$_{\pm 0.7}$ & 57.9$_{\pm 1.7}$ \\
bcGRU & 62.4$_{\pm 1.2}$ & 77.6$_{\pm 2.3}$ &66.1$_{\pm 1.7}$ &	60.5$_{\pm 1.7}$ & 84.6$_{\pm 0.5}$ & 56.9$_{\pm 2.4}$ \\
CoDE$_{mid}$ & 62.4$_{\pm 0.4}$ & 78.0$_{\pm 0.3}$ & 65.3$_{\pm 0.8}$ & 60.3$_{\pm 1.2}$ & 84.2$_{\pm 0.3}$ & 58.5$_{\pm 1.3}$\\
\hline
PT-CoDE$_{mid}$ & \bf 65.9$_{\pm 0.5}$ & \bf 81.3$_{\pm 0.4}$ & \bf 66.8$_{\pm 1.0}$ & \bf 62.6$_{\pm 0.8}$ & 84.7$_{\pm 0.3}$ & 60.4$_{\pm 0.6}$ \\
\hline
\end{tabular}
}
\begin{tablenotes}\footnotesize
\item[1]by~\cite{DBLP:conf/acl-socialnlp/LuoYC18};
\item[2]by~\cite{DBLP:conf/acl-socialnlp/Khosla18}.
\end{tablenotes}
\end{threeparttable}
\caption{Testing results on Friends and EmotionPush.}
\label{table:res_Friends_EmotionPush}
}
\hfill
%%%%%%%% Ablation on model scale
\parbox{0.34\textwidth}{
\centering
\begin{threeparttable}
\resizebox{0.34\textwidth}{!}{
\begin{tabular}{|l|l|cc|}
\hline
{\bf Model}
& {\bf Scale}
& {\bf IEMOCAP} 
& {\bf Friends} \\
\hline
\multirow{3}{*}{CoDE} 
& \textit{small} & 76.5$_{\pm 3.4}$ & \bf 62.5$_{\pm 0.7}$ \\
& \textit{mid} & \bf 78.6$_{\pm 1.3}$ & 62.4$_{\pm 0.4}$ \\
& \textit{large} & 77.6$_{\pm 1.0}$ & 62.1$_{\pm 1.0}$ \\
\cline{1-4}
\multirow{3}{*}{PT-CoDE} 
& \textit{small} & 81.2$_{\pm 0.3}$ & 65.2$_{\pm 0.4}$ \\
& \textit{mid} & \bf 81.5$_{\pm 0.4}$ & \bf 65.9$_{\pm 0.7}$ \\
& \textit{large} & 80.3$_{\pm 1.0}$  & 64.8$_{\pm 0.7}$ \\
\hline
\end{tabular}
}
\end{threeparttable}
\caption{Ablation study on model scales.}
\label{table:model_size}
}
\end{table*}

% Evaluation Metric
\subsection{ULER Evaluation Metrics}
\label{ssec:ULERmetric}

We choose F1-score~\cite{DBLP:conf/acl/TongZJM17} as the primary metric for evaluating the performance of our models. We report F1-score for each emotion considered, and use the macro-averaged F1-score of all emotion classes to evaluate the overall performance of the models, as in \cite{DBLP:conf/aaai/ZahiriC18}.
We also report the weighted accuracy (WA) and unweighted accuracy (UWA), which have been adopted in previous work~\cite{DBLP:conf/acl-socialnlp/HsuK18}. Since most of the datasets in this paper suffer the issue of imbalanced emotion distribution, F1-score is better for measuring the model performance by considering both false positives and false negatives.

% ULER Experiments
\subsection{Fine-tuning Procedure}
\label{ssec:finetune}

We still choose Adam as the optimizer and tune the learning rate for the implemented baselines. Generally, the learning rate of $2\times10^{-4}$ works well for all the datasets except MOSEI$^*$, on which we find $5\times10^{-5}$ works better. For the fine-tuning of PT-CoDE$_{mid}$, we use the learning rate of the baselines or its half and report the better results here. We monitor the macro-averaged F1-score of the validation set and decay the learning rate once the F1-score stops increasing. The decay rate and the patience of early stopping are 0.75 and 6 for all the datasets except IEMOCAP. Since IEMOCAP has very few conversations, we set the decay rate to 0.95 and the patience of early stopping to 10.

\subsection{Experimental Results}
\label{sssec:Results}

\begin{figure}[h]
    \centering
    \includegraphics[width=0.98\columnwidth]{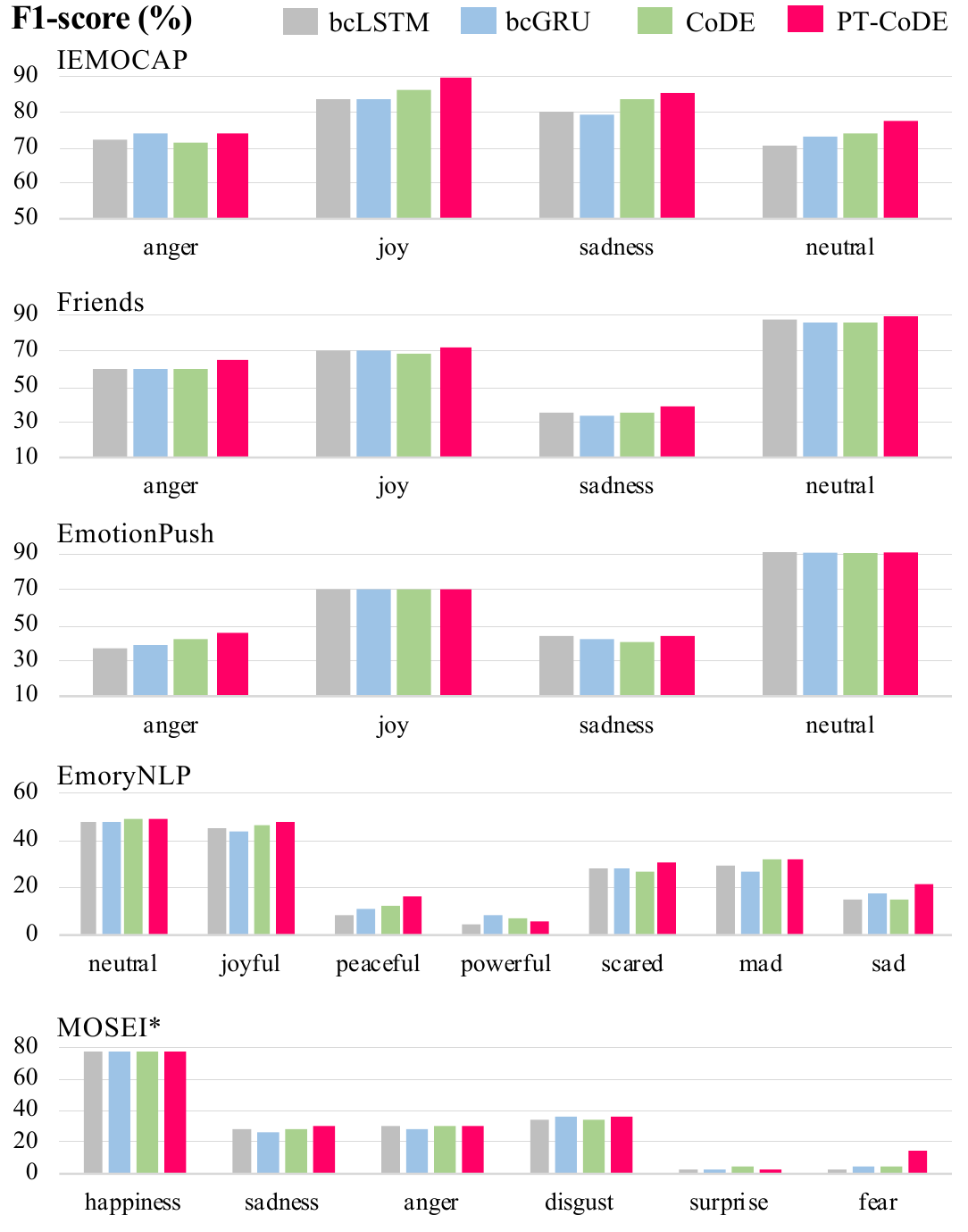}
    \caption{F1-scores for each emotion class.}
    \label{fig:F1_emotions}
\end{figure}

We report the empirical results in Table~\ref{table:res_IEMOCAP_EmoryNLP_MOSEI}, Table~\ref{table:res_Friends_EmotionPush}, and Figure~\ref{fig:F1_emotions}. The two tables present the overall performance of the models on each dataset while Figure~\ref{fig:F1_emotions} shows us the detailed F1-score distribution over the emotion classes. From these results, we have the following observations:
\\[1ex]
\textbf{Baseline Methods.}
Our implemented baselines bcLSTM$_{\ddagger}$, bcGRU, and CoDE$_{mid}$ show competitive performance as those in the previous work. 
Since the previous work~\cite{DBLP:conf/acl/PoriaCHMZM17,DBLP:conf/naacl/HazarikaPZCMZ18,DBLP:conf/acl-socialnlp/Khosla18,DBLP:conf/acl-socialnlp/LuoYC18} have not reported the F1-score on Friends, EmotionPush, and IEMOCAP, we have to compare our implemented baselines with them using the metrics of WA and UA. From the tables, we know that: {\bf(1)} On Friends, the three baselines outperform CNN-DCNN in terms of both metrics. Though SA-BiLSTM obtains a comparable result of WA, it performs much worse in terms of UA. The case of EmotionPush is a little twisted because SA-BiLSTM and CNN-DCNN perform better than the three baselines in terms of only WA or UA. But, given the good performance on Friends attained by the three baselines, which is also tested by SA-BiLSTM and CNN-DCNN, it is safe to say that the three baselines are strong ones for us to do comparative experiments.
{\bf(2)} On IEMOCAP, the three baselines outperform bcLSTM and CMN significantly when using only the textual feature, obtaining at least 3.0\% and 3.4\% absolute improvement in terms of WA and UA metrics, respectively. They even approach the performance of CMN with the trimodal features. These good results further demonstrate the strong power of our implemented baselines.
\\[1ex]
\textbf{Overall Performance.}
Our fine-tuned PT-CoDE$_{mid}$ outperforms the compared methods (previous work and implemented ones) on all the datasets in terms of F1-score. Specifically, PT-CoDE$_{mid}$ obtains at least 2.9\%, 2.8\%, 2.1\%, 2.2\%, and 2.0\% absolute improvement on IEMOCAP, Friends, EmotionPush, EmoryNLP, and MOSEI$^*$, respectively. Moreover, we find that:
{\bf(1)} On Friends, IEMOCAP, and MOSEI$^*$, our PT-CoDE$_{mid}$ beats the baselines in terms of all the three metrics. Most importantly, PT-CoDE$_{mid}$ surpasses not only the unimodal version CMN but also the trimodal version. In terms of WA and UA, we outperform the trimodal CMN by 5.3\% and 2.6\% absolute points, respectively.
{\bf(2)} On EmotionPush, our PT-CoDE$_{mid}$ achieves better result of UA, which enables us to compare it with the reported results in previous work. For SA-BiLSTM, though its WA is 3.0\% points higher than PT-CoDE$_{mid}$, PT-CoDE$_{mid}$ performs 5.4\% points better in terms of UA. Similarly, PT-CoDE$_{mid}$ performs only 2.1\% worse than CNN-DCNN in terms of UA, but 9.0\% better in terms of WA. These results indicate that our PT-CoDE$_{mid}$ benefits the minority emotion classes meanwhile does not harm the majority classes too much.
{\bf(3)} On EmoryNLP, though SCNN shows slightly better result of WA, our PT-CoDE$_{mid}$ outperforms it by 2.2\% points in terms of F1-score. It demonstrates that our PT-CoDE$_{mid}$ can better handle such an imblanced dataset.
\\[1ex]
\textbf{Specific Emotions.}
Our PT-CoDE$_{mid}$ performs as good as the implemented baselines on most emotion classes, and attains specially significant improvement on minority classes. Based on the detailed F1-scores in Figure~\ref{fig:F1_emotions}, we can summarize the primary improvement of PT-CoDE$_{mid}$ over CoDE$_{mid}$ as below:
\begin{itemize}
\item[-] IEMOCAP: \textit{anger}: +2.9\%, \textit{joy}: +3.6\%, \textit{sadness}: +1.9\%, \textit{neutral}: +3.4\%;
\item[-] Friends: \textit{anger}: +5.5\%, \textit{joy}: +3.8\%, \textit{sadness}: +2.7\%, \textit{neutral}: +2.1\%;
\item[-] EmotionPush: \textit{anger}: +4.4\%, \textit{sadness}: +3.3\%; 
\item[-] EmoryNLP: \textit{joyful}: +1.9\%, \textit{peaceful}: +3.9\%, \textit{scared}: +3.7\%, \textit{sad}: +6.5\%;
\item[-] MOSEI$^*$:  \textit{sadness}: +3.5\%, \textit{disgust}: +2.3\%, \textit{fear}: +8.1\%.
\end{itemize}
Within the summary above, \textit{anger} and \textit{sadness} are minority classes in IEMOCAP, Friends and EmotionPush, \textit{peaceful} and \textit{sad} are minority classes in EmoryNLP, and \textit{fear} and \textit{disgust} are also minority classes in MOSEI$^*$. These results demonstrate that pre-training can ameliorate the issue of imbalanced class distribution by boosting the model performance on minority classes while maintaining the good performance on majority classes.

%% F1-scores for each Emotion 

\subsection{Ablation Studies}
\label{ssec:Ablations}

Although we have demonstrated strong empirical results, we explore more about the architecture in this section. 
\\[1ex]
\textbf{Model Size.} 
We investigate how the model performance is affected by the scale of parameters. Here we consider both the CoDE and our PT-CoDE in three scales mentioned in Section~\ref{ssec:Pretraining}, i.e., \textit{small}, \textit{mid}, and \textit{large}. We test these models on IEMOCAP and Friends, and report the results in Table~\ref{table:model_size}. We find that: {\bf(1)} PT-CoDE consistently outperforms CoDE in all the three scales, demonstrating that pre-training is indeed an effective method to boost the model performance for ULER regardless of the model scale. {\bf(2)} PT-CoDE shows better performance in \textit{small} and \textit{mid} scales, we speculate that the dataset for ULER is so scarce that is incapable of transferring the pre-trained parameters of the \textit{large} scale PT-CoDE to optimal ones for this ULER task.
\\[1ex]
\textbf{Layer Effect.}
We study how different pre-trained layers affect the model performance. {CoDE+PT-U} represents that we only use the pre-trained utterance encoder but leave the conversation encoder initialized randomly. This variant obtains 1.5\% and 2.4\% absolute improvement of F1-score over CoDE, demonstrating that pre-training can result in better utterance embeddings. Comparing {CoDE+PT-U} to PT-CoDE, we can also conclude that pre-training of the conversation encoder helps the model capture the utterance-level context more effectively. {PT-CoDE+RT-W} means that we re-train PT-CoDE for 10 epochs to adjust the word embedding layer. This variant obtains only 0.1\% improvement on IEMOCAP but 1.4\% degradation on Friends, suggesting that pre-training of word embeddings does not improve the model performance necessarily but may mess up the learned utterance encoder and the conversation encoder.

\subsection{Qualitative Analysis}
\label{ssec:Cases}
We provide two examples to have a comparison between CoDE$_{mid}$ and PT-CoDE$_{mid}$. The first example comes from Friends, which are consecutive utterances from Joey. It shows that CoDE tends to recognize the utterances with exclamation marks ``!" as Angry, while those with periods ``." as Neutral. The problem also appears on PT-CoDE for short utterances, i.e., ``Push!", which contains little and misleading information. This issue might be alleviated by adding other features like audio and video. Still, PT-CoDE performs better than CoDE on longer utterances. The other example comes from EmotionPush, which are messages with few punctuation. The CoDE model predicts almost all utterances as Neutral, which may be because most of the training utterances are Neutral. However, PT-CoDE can identify the minor classes, e.g., Sad, demonstrating the strength of pre-training in alleviating the class imbalance issue.

% Layers
\begin{table}[t!]
\small
\centering
\begin{threeparttable}
%\resizebox{0.98\columnwidth}{!}{
\begin{tabular}{|l|cc|}
\hline
{\bf Layers}
& {\bf IEMOCAP} 
& {\bf Friends} \\
\hline
PT-CoDE + RT-W & \bf 81.6$_{\pm 0.5}$ & 64.5$_{\pm 0.9}$ \\
PT-CoDE & 81.5$_{\pm 0.4}$ & \bf 65.9$_{\pm 0.7}$ \\
CoDE + PT-U & 80.1$_{\pm 1.3}$ & 64.8$_{\pm 0.7}$ \\
CoDE & 78.6$_{\pm 1.3}$ & 62.4$_{\pm 0.4}$ \\
\hline
\end{tabular}
%}
\end{threeparttable}
\caption{Ablation study on pre-trained layers.}
\label{table:layers}
\end{table}

% Cases
\begin{table}[t!]
\fontsize{10}{11}\selectfont
    \centering
    \resizebox{\columnwidth}{!}{
    \begin{tabular}{|c|p{4cm}|c|c|c|}
    \hline
        \bf Speaker & \quad\quad\quad \bf Utterance & \bf Truth & \bf CoDE & \bf PT-CoDE \\
    \hline
    \hline
        \bf Example 1 & & & & \\
    \hline
        Joey & Come on, Lydia, you can do it.				 & Neu & Neu & Neu\\
        Joey & Push!							             & Joy & \colorbox{red!30}{Ang} & \colorbox{red!30}{Ang}\\
        Joey & Push 'em out, push 'em out, harder, harder.   & Joy & \colorbox{red!30}{Neu} & \colorbox{red!30}{Neu}\\
        Joey & Push 'em out, push 'em out, way out!			 & Joy & \colorbox{red!30}{Ang} & Joy\\
        Joey & Let's get that ball and really move, hey, hey, ho, ho.	 & Joy & \colorbox{red!30}{Neu} & Joy\\
        Joey & Let's…  I was just… yeah, right.				 & Joy & \colorbox{red!30}{Neu} & \colorbox{red!30}{Neu}\\
        Joey & Push!							             & Joy & \colorbox{red!30}{Ang} & \colorbox{red!30}{Ang}\\
        Joey & Push!							             & Joy & \colorbox{red!30}{Ang} & \colorbox{red!30}{Ang}\\
    \hline
    \hline
        \bf Example 2 & & & & \\
    \hline
        Sp1 & It's so hard not to cry					     & Sad & \colorbox{red!30}{Ang} & Sad\\
        Sp2 & What happened						             & Neu & Neu & Neu\\
        Sp1 & I lost another 3 set game					     & Sad & \colorbox{red!30}{Neu} & Sad\\
        Sp2 & It’s ok person\_145					         & Neu & Neu & Neu\\
        Sp1 & Why does it hurt so much				         & Sad & \colorbox{red!30}{Neu} & Sad\\
        Sp2 & Everybody loses					             & Neu & Neu & Neu\\
    \hline
    \end{tabular}
    }
    \caption{\label{table:cases} Qualitative comparison between CoDE$_{mid}$ and PT-CoDE$_{mid}$ by two examples.
    }
\end{table}

\section{Conclusion}
\label{sec:conclusion}

We propose to pre-train a context-dependent encoder (CoDE) for the utterance-level emotion recognition (ULER) task. The CoDE is essentially a hierarchical architecture, which differs from existing works that aim to pre-train a universal sentence encoder. We also propose the CoCo task, i.e., the ``conversation completion" task, to pre-train CoDE on movie subtitles in an unsupervised fashion. Finally, the pre-trained CoDE (PT-CoDE) is fine-tuned for ULER and significantly boosts the model performance on five datasets for ULER.
A possible future direction is to conduct two-step pre-training, i.e., to utilize the prevalent pre-trained models like GPT and BERT as the utterance encoder and then pre-train the conversation encoder by ourselves.
Another is to incorporate the word-level self-supervision information into pre-training to further improve utterance features.

% references
%\newpage
\bibliography{naaclhlt2019}
\bibliographystyle{acl_natbib}

\appendix

\end{document}